\documentclass[conf]{new-aiaa}
\usepackage[utf8]{inputenc}

\usepackage{graphicx}
\usepackage{amsmath}
\usepackage[version=4]{mhchem}
\usepackage{siunitx}
\usepackage{longtable,tabularx}
\usepackage{epsfig}
\usepackage{hyperref}
\usepackage{color}
\usepackage[normalem]{ulem}

\usepackage{caption}
\usepackage{subcaption}

\usepackage{algorithm}
\usepackage{algpseudocode}

\title{Learning to Communicate: A Machine Learning Framework for Heterogeneous Multi-Agent Robotic Systems}

\author{Hyung-Jin Yoon\footnote{Graduate Student, Mechanical Engineering, University of Illinois at Urbana-Champaign},
Huaiyu Chen\footnote{Undergraduate Student, Mechanical Engineering, University of Illinois at Urbana-Champaign},
Kehan Long\footnote{Undergraduate Student, Department of Mathematics, University of Illinois at Urbana-Champaign}, 
Heling Zhang\footnote{Undergraduate Student, Electrical \& Computer Engineering, University of Illinois at Urbana-Champaign},

Aditya Gahlawat\footnote{Postdoctoral Research Associate, Mechanical Engineering, University of Illinois at Urbana-Champaign},
Donghwan Lee\footnote{Postdoctoral Research Associate, Coordinated Science Laboratory, University of Illinois at Urbana-Champaign}, and Naira Hovakimyan\footnote{Professor, Mechanical Engineering, University of Illinois at Urbana-Champaign AIAA Fellow.}}

\begin{document}

\maketitle

\begin{abstract}
We present a machine learning framework for multi-agent systems to learn both the optimal policy for maximizing the rewards and the encoding of the high dimensional visual observation. The encoding is useful for sharing local visual observations with other agents under communication resource constraints. The actor-encoder encodes the raw images and chooses an action based on local observations and messages sent by the other agents. The machine learning agent generates not only an actuator command to the physical device, but also a communication message to the other agents. We formulate a reinforcement learning problem, which extends the action space to consider the communication action as well. The feasibility of the reinforcement learning framework is demonstrated using a 3D simulation environment with two collaborating agents. The environment provides realistic visual observations to be used and shared between the two agents.
\end{abstract}

\section{Introduction}
Communication is crucial for the satisfactory performance of multi-agent systems. Different sensors and actuators of the agents can be better used when their individually collected information is shared and collaboratively processed. However, designing communication protocols suitable for multi-agent systems is not a trivial task. The low-cost, high-resolution image sensors can provide a large amount of information, which might be hard to process in real time. Moreover, the transmission of information is constrained by the limited bandwidth of the communication network. It is therefore desirable for the communication protocol to compress the visual data to allow its transmission over the resource-constrained network, provided that the vital for the collaborative mission execution is not lost. In light of these considerations, we propose a machine learning framework for multi-agent systems, where visual information is shared between the agents to accomplish collaborative tasks. The proposed framework employs a reinforcement learning problem formulation. The control policy of each agent dictates not only the local actuator commands, but also the communication messages for transmission to the other agents. The approach is tested in a 3D simulation environment, developed using a game/virtual reality development tool.  As an experimental validation, we implement the proposed algorithm for collaborative search in a game environment using a team of a high-altitude and low-altitude aerial vehicles. 

It is important to mention that application of the proposed approach to real-world environment for cooperative flight missions, as the one in~\cite{kaminer2017time}, should be considered only after the {\em safety} issues of the end-to-end deep reinforcement learning (DRL) are addressed. Training the deep neural network with a large number of parameters requires an even larger number of training samples. Typically, training of DRL takes a few millions of time-steps~\cite{mnih2015human, lillicrap2015continuous}, which is not affordable in real flight missions. An incidence of catastrophic failure is very likely during the transient of this long training period. This fundamental drawback has prevented safety-critical applications of deep learning methods and the variants of those. However, the capability to process high-dimensional sensor signals, such as camera images, have attracted a broad audience to develop vision-based control applications. Some of the evident examples in safety-critical applications are collision avoidance maneuvers that  cannot afford trials and errors by DRL. A possible solution to this issue is to restrict the role of DRL in the optimization, while ensuring the safety during the transient by model-based controllers~\cite{shalev2016safe}. We note that {\em safe reinforcement learning} is not in the scope of this paper. Instead, we focus on the communication between the agents and the centralized policy improvement module (central critic) within a cooperative decision making framework. Our work of exploring the potential of DRL for multi-agent systems is a small step to bring DRL from experimental science to practical engineering.

\subsection{Related Work}
Reinforcement learning has been applied to various multi-agent systems~\cite{busoniu2008comprehensive, matignon2012independent, lowe2017multi, foerster2017counterfactual}. Since the agents are co-evolving together, traditional reinforcement learning approaches such as Q-learning~\cite{watkins1992q} do not perform well as they do with single-agent applications. For example, independent reinforcement learning formulations for multi-agent systems, where agents do not exchange any information and treat the other agents as a part of the environment, have displayed poor performance~\cite{matignon2012independent}. A possible reason for this could be that the co-evolving of the other agents breaks the Markov assumption of the Markov decision process, which is the typical model for most reinforcement learning techniques.

The framework of \emph{centralized training with decentralized execution} has been successfully applied to multi-agent reinforcement learning problems~\cite{lowe2017multi, foerster2017counterfactual}. Since this framework allows a central critic to access global information generated by the agents, the critic can still rely on the Markov assumption to estimate policy evaluation using the Bellman equation~\cite{sutton1998reinforcement}. However, this framework relies on the communication of all observations and actions from all local agents for the centralized training despite formulating distributed control policies. In the case of visual observations, sending high-resolution images from many locally distributed agents over a wireless network can be costly or infeasible.

Information sharing between agents in the context of reinforcement learning has been explored in~\cite{foerster2016learning, mao2017accnet}. The authors of the cited papers demonstrate that the reinforcement learning problem formulation can be useful to find communication policies (protocols). This is accomplished by including performance metrics (to be optimized) in the rewards of the underlying Markov decision process. In a similar fashion, we use the reward function to ensure that the messages capture enough information for the centralized critic to evaluate local policies.

In this paper, we extend the framework of \emph{centralized training with decentralized execution} to include  additional optimization of the inter-agent communication. Specifically, we employ the multi-agent deep deterministic policy gradient (MADDPG) algorithm, introduced in~\cite{lowe2017multi}, and extend the action space to consider communication between the agents and the central critic. Also, we consider autoencoders~\cite{hinton2006reducing} to compress high dimensional visual observations into low dimensional features. This allows each agent to send local visual observations to the central critic, while not violating the constraints of communication resource usage.

The remainder of the paper is organized as follows. In Section II, the proposed reinforcement learning method that considers communication actions for a multi-agent system is presented. In Section III, we introduce a 3D simulation environment and implement the proposed reinforcement learning method in the simulation environment. In Section IV, the simulation results are analyzed. Section V summarizes the paper.

\section{Method}\label{sec:method}
We begin by formulating the Markov decision process (MDP) for multi-agent systems. The MDP for $N$ agents is defined by a set of all possible states $\mathcal{S}$ of the global environment, a set of actions $\mathcal{A}_1, \dots, \mathcal{A}_N$, and a set of local observations of the environment $\mathcal{O}_1, \dots, \mathcal{O}_N$. The i\textsuperscript{th} agent\footnote{We use superscript $(i)$ to denote the $i$\textsuperscript{th} agent.}, $i \in \{1,\dots,N\}$, chooses its actions from $\mathcal{A}_i$ and receives observations from $\mathcal{O}_i$. Furthermore, given a local observation, the $i\textsuperscript{th}$ agent chooses its action via a deterministic policy  $\mu^{(i)}_{\theta^{(i)}}:\mathcal{O}_i \rightarrow \mathcal{A}_i$, which is parameterized by the real-valued vector $\theta^{(i)}$. For the complete system of $N$-agents, we define the joint-policy $\mu_\theta: {\cal O} \to {\cal A}$ as $\mu_\theta:=(\mu^{(1)}_{\theta^{(1)}},\mu^{(2)}_{\theta^{(2)} },\ldots,\mu^{(N)}_{\theta^{(N)}})$, where $\theta:=(\theta^{(1)},\theta^{(2)},\ldots,\theta^{(N)})$, ${\cal O}:={\cal O}_1 \times {\cal O}_2 \times \cdots \times {\cal O}_N$, and ${\cal A}:={\cal A}_1 \times {\cal A}_2 \times \cdots\times {\cal A}_N$. At any given time $t \in \mathbb{N}$, we denote the action taken by Agent $i$ by $\mathbf{a}^{(i)}_t$. The time-dependent action $\mathbf{a}^{(i)}_t$ is defined as 
\begin{equation}\label{eqn:action}
    \mathbf{a}_t^{(i)}:= (a^{(i)}_{m,t}, a^{(i)}_{c,t}) \in {\cal A}_i,
\end{equation} where $a^{(i)}_{m,t}$ is the physical action taken by Agent $i$ through its actuators, and $a^{(i)}_{c,t}$ is a virtual action to be communicated to the other agents. As previously described, Agent $i$ uses its local policy $\mu^{(i)}_{\theta^{(i)}}$ to generate actions $\mathbf{a}^{(i)}_t$ at each $t$, thus, we may write
\[\mathbf{a}_t^{(i)}:= (a^{(i)}_{m,t}, a^{(i)}_{c,t}) = \mu^{(i)}_{\theta^{(i)}}(o^{(i)}_t), \quad o^{(i)}_t \in \mathcal{O}_i.\]

The state-transition model $\mathcal{T}$ determines the probability of the next global state given the current state and the actions taken by all the agents, i.e., $\mathcal{T}:\mathcal{S}\times\mathcal{A}_1\times\dots\times\mathcal{A}_N\times\mathcal{S}'\rightarrow[0,1]$. Since we consider the operation of a team of agents, and not individual agents, we consider collaborative rewards. To be precise, we consider the collaborative reward $r$, which depends on the global states, the actions, and the next state, i.e.  $r:\mathcal{S}\times\mathcal{A}_1\times\dots\times\mathcal{A}_N\times\mathcal{S}'\rightarrow \mathbb{R}$. Therefore, the goal is to maximize the expectation of the sum of all future rewards $\sum_{t=0}^{\infty}\gamma^{t} r_t$, where $\gamma \in (0,1)$ is a discount factor, and $r_t$ is the reward earned at time step $t$.   

\begin{figure}[h!]
\centering
 \includegraphics[width=0.8\textwidth]{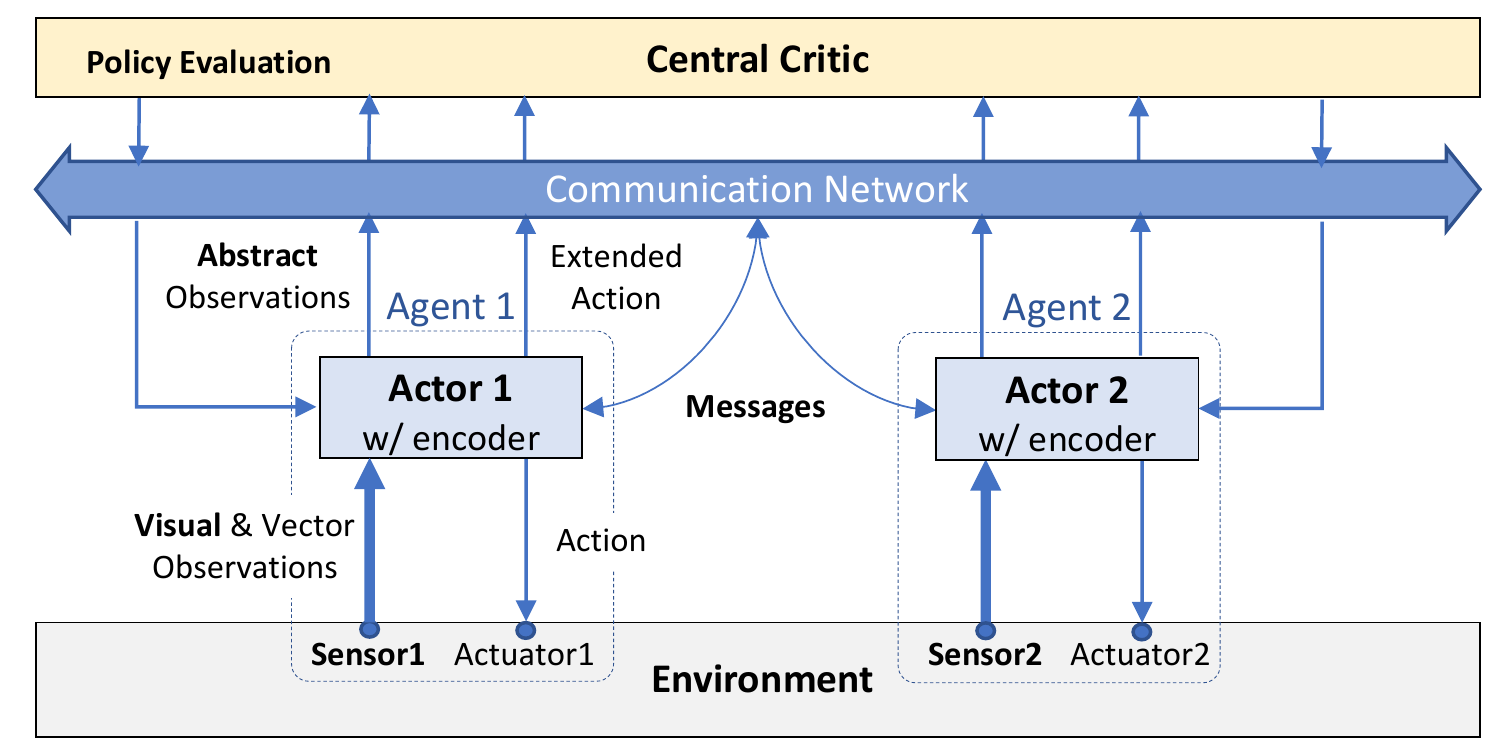}
 \caption{Overview of decentralized policy (actor) and central critic.}
\medskip
\label{fig: Diagram}
\end{figure}

We consider a two-agent system, presented in Figure~\ref{fig: Diagram}, as a concrete example of the proposed framework. Both Agents~$1$ and~$2$ have onboard actors that compute their  local control policies $\mu_{\theta^{(i)}}^{(i)}$, $i \in \{1,2\}$. Additionally, using the communication network, each agent interacts with a centralized critic, which is an algorithm that evaluates the performance (value functions) of the agents. Further details on the actor-critic framework can be found in~\cite{konda2000actor}. Each actor determines the action taken by its agent using the signals from onboard sensors like cameras and IMUs, and the virtual action $a_{c,t}^{(i)}$ of the other agent (see Eqn.~\eqref{eqn:action}). We refer to the virtual actions shared between the agents as messages in Figure~\ref{fig: Diagram}. In addition to the actions taken by the agents, the central critic requires \textit{all} the observations available to each agent in order to evaluate the performance of the multi-agent system. 
The sensor measurements, actions, and messages (virtual actions) can be communicated easily. However, the communication of high-resolution raw images recorded by the onboard cameras of each agent creates a considerable network burden and thus presents a significant challenge. In order to mitigate this issue, each agent is also equipped with an autoencoder which transforms the raw camera images into low dimensional features (encoded images). Then, the sensor measurements, messages, actions, and low dimensional features, which we collectively refer to as abstract observations, can be transmitted with ease between the agents and the central critic. This is due to the fact that the encoder compresses the raw images significantly. To summarize, the agents communicate their respective virtual actions with each other while transmitting their abstract observations to the central critic. The critic then communicates the performance evaluation back to each agent. 

For the considered multi-agent setup, we need to address the two major components: i) the central critic, and ii) the autoencoders and actors onboard each agent. We refer to the actor and the autoencoder jointly as the actor-encoder. 
% Message optimization refers to the choice of messages (virtual actions $a_{c,t}^{(i)}$ in~\eqref{eqn:action}) which the agents communicate with each other. This optimization is automatically handled by the reinforcement learning framework via the inclusion of the global action (action taken by all the agents) in the collaborative reward function $r_t$ which we previously defined to be
% \[r_t = r(s_t,\mathbf{a}_t),\] and where $s_t$ is the global state. Since, by~\eqref{eqn:action},
% \[\mathbf{a}_t=(\mathbf{a}_t^{(t)},\mathbf{a}_t^{(2)}),\quad \text{and} \quad \mathbf{a}_t^{(i)}:= (a^{(i)}_{m,t}, a^{(i)}_{c,t}),\] we see that the message/virtual actions are encoded in the collaborative reward $r_t$.

\noindent \textbf{Central Critic:}
Reinforcement learning algorithms developed for single agents can be applied to multi-agent systems by using the \emph{centralized training with decentralized execution (CTDE)} framework, see, for e.g.~\cite{lowe2017multi, foerster2017counterfactual}. Similar to the work in~\cite{lowe2017multi}, we apply the CTDE framework to the deep deterministic policy gradient (DDPG) algorithm~\cite{lillicrap2015continuous}, which is a variant of the deterministic policy gradient (DPG)~\cite{silver2014deterministic}.  In the DDPG algorithm, the critic estimates the true state-action value function $q^{\mu_\theta}(s, a)$ for the joint policy $\mu_\theta$ using the following recursive equation:
\begin{equation}\label{eq: true_q_function}
    q^{\mu_\theta}(s_t, \mathbf{a}_t) = {\mathbb E}[r(s_t, \mathbf{a}_t) + \gamma q^{\mu_\theta}(s_{t+1}, \mathbf{a}_{t+1})],
\end{equation}
where the expectation ${\mathbb E}$ is taken with respect to the state-action distribution determined by the policy $\mu$ and the underlying MDP, and $\gamma \in (0,1)$ is the discount factor. The critic cannot directly use~\eqref{eq: true_q_function} to estimate $q^{\mu_\theta}(s, a)$, since it does not have access to the true global state $s_t$. Instead, the proposed critic uses the abstract observations transmitted from each agent to estimate the state-action value function. The critic approximates the state-action value function $q^{\mu_\theta}(s, a)$ by $Q(\mathbf{o},\mathbf{a};w)$, where $w$ is a parameter, and $\mathbf{a}$ is the global action (actions taken by all the agents), which, using~\eqref{eqn:action}, can be expressed as
\[
\mathbf{a}=(\mathbf{a}^{(1)},\mathbf{a}^{(2)} ),\quad \mathbf{a}^{(i)} = (a^{(i)}_{m}, a^{(i)}_{c}), \quad i \in \{1,2\}. 
\] Note that we have dropped the time-dependence of $\mathbf{a}$ on $t$ for brevity. Additionally, $\mathbf{o}$ is the concatenation of all the observations communicated by all the agents to the critic. Therefore, $\mathbf{o}$ contains the sensor measurements collected by each agent, the messages containing the virtual actions transmitted between the agents, and the abstract visual observations (encoded images) compressed by the autoencoders onboard each agent. The parameter $w$ in $Q(\mathbf{o},\mathbf{a};w)$ is estimated by minimizing the temporal difference (TD) error~\cite{tsitsiklis1997analysis}. The loss function $l(w)$ to be minimized for  the estimation of $Q(\mathbf{o}, \mathbf{a};w)$ is  
\begin{equation}\label{eq:loss_critic}
l(w)=\frac{1}{M}\sum_{m=1}^M(Q(\mathbf{o}_m, \mathbf{a}_m;w)-y_m)^2, \quad y_m = r_m + \gamma Q(\mathbf{o}'_{m}, \mu_\theta(\mathbf{o}'_m);w),
\end{equation}
where $M$ transitions are uniform random samples from a replay buffer $\mathcal{D}_\text{replay}$, which stores the recently observed $L$ transitions, i.e.:
\begin{equation*}
 \mathcal{D}_{\text{replay}}:=(\,(\mathbf{o}_{t-L}, \mathbf{a}_{t-L}, r_{t-L}, {\mathbf{o}'}_{t-L})\,, \dots\,, (\mathbf{o}_{t-1}, \mathbf{a}_{t-1}, r_{t-1}, {\mathbf{o}'}_{t-1})\,).
\end{equation*}
Here, each transition $(\mathbf{o}, \mathbf{a}, r, {\mathbf{o}'})$ consists of the concatenated abstract observation $\mathbf{o}$, the global action by all agents $\mathbf{a}$, collaborative reward $r$, and the next observation $\mathbf{o}'$. Our use of the replay buffer $\mathcal{D}_\text{replay}$ is motivated by its successful application to \emph{Deep Reinforcement Learning}~\cite{mnih2015human, lillicrap2015continuous}. However, the size of replay buffers used in \cite{mnih2015human, lillicrap2015continuous} is  million samples. For the type of multi-agent systems that we consider, which use high-resolution images, the storage of such replay buffers would be infeasible\footnote{A million color images with $200 \times 200$ resolution would take 480 Gigabyte of memory to store.}. This is where our novel approach of encoding raw images is advantageous, since the compressed images require significantly less memory resource to store. Thus, our approach of encoding high-resolution images not only enables low-burden communication, but also the storage of replay buffers, which is crucial for the operation of the central critic.  

\noindent \textbf{Autoencoders and Actors:} We now explain the autoencoders and actors onboard each agent. As aforementioned, the central critic requires the concatenated observation $\mathbf{o}$, a major component of which is the set of compressed images recorded by each agent's onboard camera. This compression is performed by autoencoders~\cite{hinton2006reducing} onboard each agent. The efficacy of using autoencoders in reinforcement learning was empirically demonstrated in~\cite{lange2012autonomous}. In the proposed approach, the autoencoders are learned on-line. The autoencoder consists of an encoder and a decoder. To ensure that the encoder does not remove the principle components from the raw image, an image is reconstructed by the decoder from the encoded image and compared to the original raw image. This comparison then guides the learning of the encoder. We denote the encoder onboard Agent $i$ by $f(\cdot;\xi_{e}^{(i)})$, where $\xi_{e}^{(i)}$ are the parameters to be learned. Given a set of images $\mathcal{I}$, the encoder compresses the images to a pre-specified dimension $D$, i.e. $f(\cdot;\xi_{e}^{(i)}):\mathcal{I} \rightarrow \mathbb{R}^D$. The decoder, denoted by $g(\cdot;\xi_{d}^{(i)}):\mathbb{R}^D \rightarrow\mathcal{I}$, performs the reconstruction. Here, $\xi_{d}^{(i)}$ are the parameters of the decoder. The encoder and decoder parameters, $\xi_{e}^{(i)}$ and $\xi_{d}^{(i)}$, respectively, are learned by recursively minimizing the loss: 
\begin{equation}\label{eq: encoder_loss}
    l(\xi_{e}^{(i)},\xi_{d}^{(i)}) = \frac{1}{P}\sum_{p=1}^P d(v_p^{(i)}, g(f(v_p^{(i)};\xi_{e}^{(i)});\xi_{d}^{(i)})),
\end{equation}
where $d(\cdot, \cdot)$ is a difference metric\footnote{Mean squared error is an example of metric between images.} between two images. Additionally, $\{v_p^{(i)}\}_{p=1}^P$ is the mini-batch sample from image data buffer $\mathcal{D}_{\text{image}}^{(i)}$ stored locally on the $i^{\text{th}}$ agent and contains the recently observed camera images taken by the agent. Upon completion of the compression, each agent communicates its encoded image $f(v_p^{(i)};\xi_{e}^{(i)})$ to the central critic, which is then used to construct the concatenated observation $\mathbf{o}$. 

Once the central critic computes the approximate state-action value function $Q(\mathbf{o},\mathbf{a};w)$ using~\eqref{eq:loss_critic}, it communicates this value and the associated mini-batch back to each agent. The onboard actors use these to improve their local policies $\mu_{\theta^{(i)}}^{(i)}$, $i \in \{1,2\}$. As previously mentioned, the actors use deep deterministic policy gradient (DDPG) algorithm~\cite{lillicrap2015continuous} to improve the policies of their agents by estimating the policy gradient based on the communicated mini-batch as
\begin{equation}
    \nabla_{\theta^{(i)}}J     \approx  \frac{1}{P}\sum_{p=1}^P \nabla_{\mathbf{a^{(i)}}}Q(\mathbf{o},\mathbf{a^{(i)}};w)|_{\mathbf{o}=\mathbf{o}_p, \mathbf{a^{(i)}}=\mu_{\theta^{(i)}}^{(i)}(\mathbf{o}_p)}  \nabla_{\theta^{(i)}} \mu_{\theta^{(i)}}^{(i)} (\mathbf{o})|_{\mathbf{o}_p}.
\end{equation} Here, $\mathbf{o}_p$ is the concatenated observation contained in the communicated mini-batch.
Then, each agent's local policy is improved by recursively adding the policy gradient $\nabla_{\theta^{(i)}}J $ to the current policy $\mu_{\theta^{(i)}}^{(i)}$ with a decaying gain (step-size) in order to improve the expected sum of the future rewards.
 Finally, we would like to mention the process by which each agent optimizes the messages (virtual actions $a_{c,t}^{(i)}$ in~\eqref{eqn:action}) to communicate with other agents. This optimization is automatically handled by the reinforcement learning framework via the inclusion of the global action (action taken by all the agents) in the collaborative reward function $r_t$, which we previously defined to be
 \[r_t = r(s_t,\mathbf{a}_t),\] and where $s_t$ is the global state. Since, by~\eqref{eqn:action},
 \[\mathbf{a}_t=(\mathbf{a}_t^{(1)},\mathbf{a}_t^{(2)}),\quad \text{and} \quad \mathbf{a}_t^{(i)}:= (a^{(i)}_{m,t}, a^{(i)}_{c,t}),\] we see that the message/virtual actions are encoded in the collaborative reward $r_t$.
 
In conclusion, the proposed reinforcement learning algorithm has 3 parameter updates: (1) autoencoder, (2) critic, and (3) policy update. Since the algorithm has multiple parameters being updated simultaneously, we require that each parameter update uses different time-scales for the associated step-sizes. This is because, as explained in~\cite{konda2000actor}, the policy needs to be updated slowly so that the critic can track the changes of the Markov chain (Controlled MDP). Similarly, we employ the approach of the multiple-time-scale algorithm~\cite{kushner2003stochastic}. The autoencoder is updated at the slowest time-scale, and the critic is updated at the fastest time-scale. This allows the policy evaluation performed by the critic to track the slowly varying changes of the Markov chain.

\section{Experiments}
We develop a 3D simulation environment to test and validate the proposed algorithms for the two-agent system illustrated in Figure~\ref{fig: Diagram}. The 3D environment simulates an urban scene, in which two unmanned aerial vehicles (UAVs) operate collaboratively to identify and approach a person of interest. The two UAVs represent the agents in our framework. The first agent flies at a low-altitude and is equipped with a front-view camera and sensors measuring its position and velocity. The second agent operates at a higher altitude and is equipped with a down-facing camera. The first agent's front-view camera images are not sufficient to allow it to search and move towards the target person. Therefore, it must also rely on the down-facing camera images taken by the second agent. Figure~\ref{fig: RL_Envrionment} illustrates the positions of the agents in the simulation environment.

We notice that similar problems have been considered in cooperative path-following tasks in \cite{xargay2012time, kaminer2017time}, where the two heterogeneous UAVs had to execute a cooperative road search mission. Here we focus on decision making for strategy development through sharing of information, while in \cite{xargay2012time, kaminer2017time} the focus was on cooperative path following, where the paths were generated apriori. The framework of this paper can lead to (near) real-time optimal navigation algorithms that a cooperative path following framework can use efficiently.

\subsection{The 3D Simulation Environment}\label{subsec:sim}
We now explain the development of the simulator and the setup of the experiments. We use \emph{Unity 3D}~\cite{juliani2018unity}, a game development editor, to construct the environment. Using the physics engine of the editor, we can conveniently model the rigid-body dynamics of the agents. The physics engine updates the state of the rigid body dynamics at 60 Hz in the game time\footnote{The game engine keeps track of the states of the environment with its own timestamps that we refer to as game time.}. A proportional–integral–derivative (PID) controller generates the moment and the force exerted on the agent to track the velocity commands given by the onboard actors. The first agent can execute forward, backward, and yaw movements, while the high-altitude second agent hovers at a fixed position. The proposed algorithm interacts with the game environment at 10 Hz in game time. Furthermore, the game engine checks for collisions of the agents with all objects in the scene such as buildings, cars, and traffic signs. Such a collision results in a negative reward for the respective agent, thus encouraging a collision-free safe behavior. Additionally, an important feature of our simulator is the realistic visualization of the 3D environment. This allows us to simulate the capture of high-resolution images by cameras onboard each agent. 

\begin{figure}[thpb]
\centering
 \includegraphics[width=0.8\textwidth]{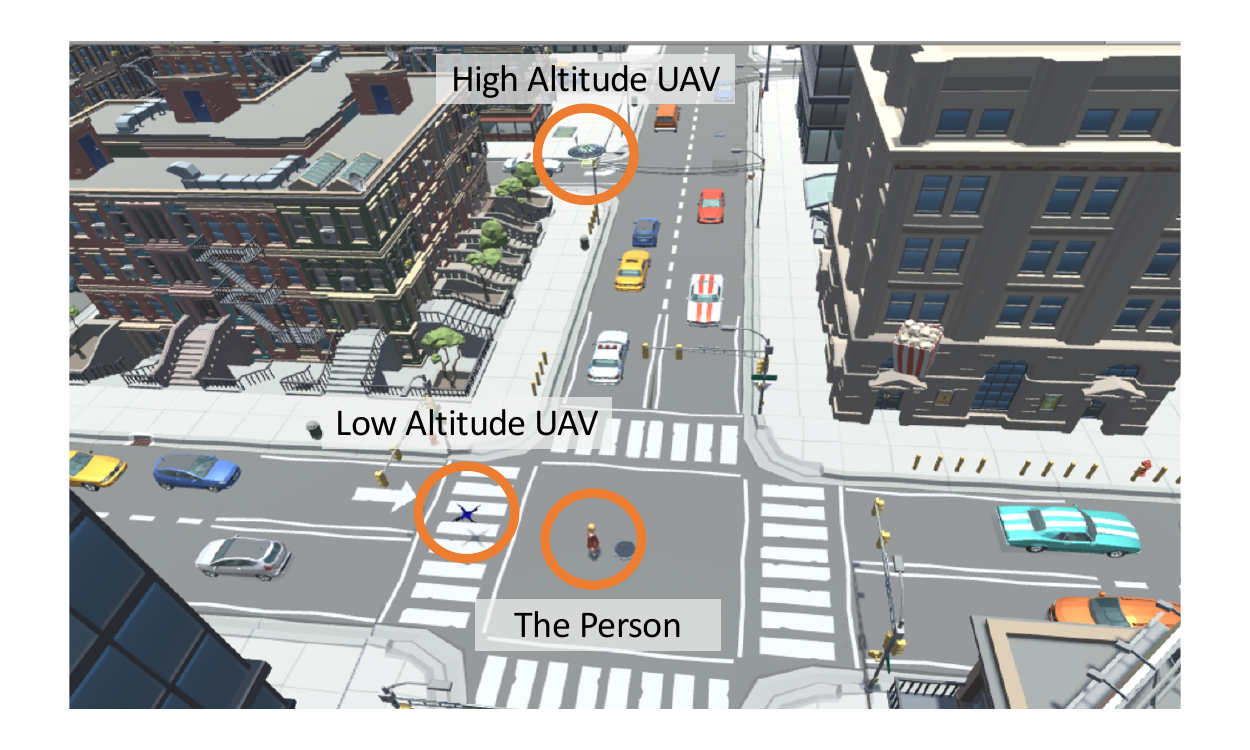}
 \caption{The 3D Environment with two agents: the first agent (low-altitude UAV)'s task is to reach the person, and the second agent (high-altitude UAV) sends messages to the first agent to share the top-view. An illustrative video of the environment can be found~{[\href{https://youtu.be/gNsxl0RGhdA}{here}]}.}  
\medskip
\label{fig: RL_Envrionment}
\end{figure}
\begin{figure}[thp]
\centering
\begin{subfigure}{.5\textwidth}
  \centering
  \includegraphics[width=1.\linewidth]{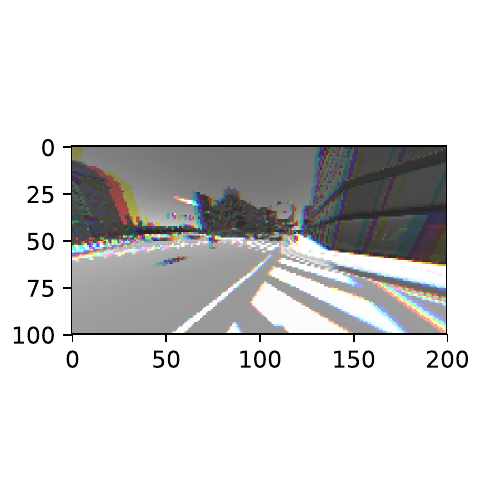}
  \caption{Stacked 3 frames of front view by the low-altitude UAV.}
  \label{fig: image1}
\end{subfigure}%
\begin{subfigure}{.5\textwidth}
  \centering
  \includegraphics[width=1.\linewidth]{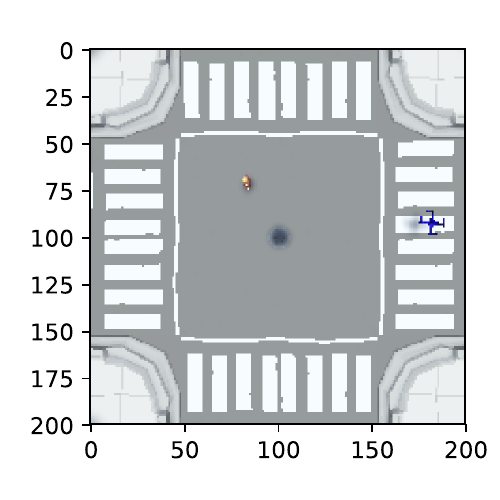}
  \caption{Color image of top view by the high-altitude UAV.}
  \label{fig: image2}
\end{subfigure}
\caption{The high dimensional visual observations: (1) The stacked frames image (size: $3\times100\times200$) contains the images taken in the recent three frames; (2) The color image (size: $3\times200\times200$) has information of the person's location.}
\label{fig: likelihood_and_sigma}
\end{figure}

To train the various components of the proposed algorithm (central critic, onboard actors, and autoencoders), we run episodes of the collaborative search task in the simulator. As mentioned before, the two-agent team is tasked with finding and approaching the target person. The first agent (low-altitude UAV) and the second agent (high-altitude UAV) must communicate with each other and the central critic so that they can efficiently use each other's different observations to accomplish the task. Each episode is designed as follows:

\begin{enumerate}
    \item \textbf{Episode initialization:} Both the target person and the first agent start at uniformly random positions and orientations. These initial positions always lie on the intersection in Figure~\ref{fig: RL_Envrionment}. The second agent is initialized at a fixed position overlooking the intersection.
    
    \item \textbf{Episode termination:} An episode is terminated, when the two-agent team either succeeds or fails. The team succeeds, if the first agent gets within 3 meters of the target person. Conversely, the team fails, if the first agent either gets more than 30 meters away from the target person or collides with an object.
    
    \item \textbf{Reward design:} In order to train the central critic and the agents' actors and autoencoders, we design the rewards as follows: the successful completion of the task leads to a reward of 100, while a failure leads to a reward of -100. Furthermore, the central critic receives a reward of 1 for every time step, over which the first agent moves closer to the target person. Finally, we impose a reward of -0.05 per time step as a time penalty.  
\end{enumerate}

During each episode, the two-agent team operates using the method described in Section~\ref{sec:method}. Both agents communicate messages (virtual actions) between each other. Additionally, each agent uses its autoencoder to compress its camera images and communicates the abstract observations (sensor measurements, compressed images, and actions) to the central critic. The central critic uses these inputs from the agents to approximate the state-action values and communicates it back to each agent along with the associated mini-batches. Each agent's actor then uses this information to improve its policy and take actions. This process repeats at each time step. The design of rewards ensures that over multiple episodes the two-agent team learns to collaborate and ensure that the task is completed, i.e., the first agent successfully approaches the target person.

\subsection{Implementation of Multi-Agent Deep Deterministic Policy Gradient (MADDPG)}
Algorithm~\ref{alg1} describes the proposed multi-agent reinforcement learning framework. Compared to the MADDPG algorithm~\cite{lowe2017multi}, the proposed framework has an additional component, an autoencoder that encodes high dimensional images into low dimensional features to be used by the central critic. As mentioned previously, with this extension, we are able to keep the memory resource usage for replay buffer within a reasonable range.

\begin{figure}[thp]
\centering
 \includegraphics[width=1.0\textwidth]{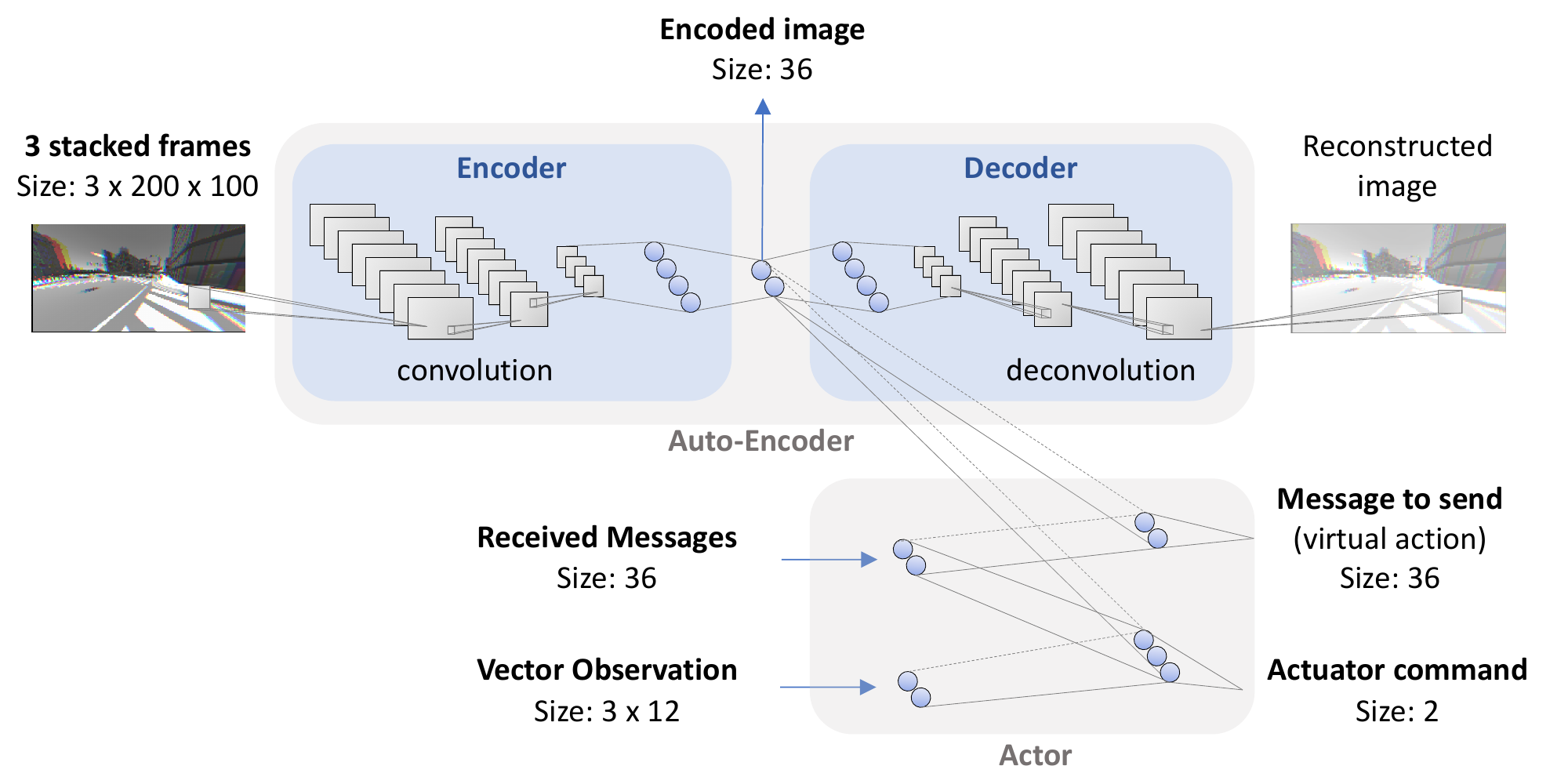}
 \caption{The computation graph of the actor-encoder network for the first agent. The convolution layers are shared by the autoencoder and the actor. The vector observation is the stacked 3 frames of 12-dimensional state of the rigid body. Batch normalization~\cite{ioffe2015batch} is used between layers to keep the signals within an appropriate scale. The second agent uses the same neural network without generating actuator command.}
\medskip
\label{fig: actor-encoder}
\end{figure}

We apply the (MADDPG)~\cite{lowe2017multi} algorithm to the two agent system explained in Section~\ref{subsec:sim}. Each agent has a different set of sensors and actuators. 
The first agent senses its 12-dimensional rigid body state, the stacked frames of the front-view camera mounted on it as shown in Figure~\ref{fig: image1}, and receives messages from the second agent. The first agent is actuated by a PID controller, which receives forward, backward, and yaw commands from its onboard actor. The second agent observes the color images of the top view as shown in Figure~\ref{fig: image2} and the messages sent by the other agent. The second agent does not have any actuators. The second agent's action contains only communication messages to inform the first agent of the location of the person inferred from the top view images. The communication messages sent at the current time step become available to the other agents in the next time step. The communication messages contain the virtual actions generated by the onboard actors in Figure~\ref{fig: actor-encoder}.  

As mentioned earlier, the performance of the agents is evaluated by the central critic. The central critic uses the concatenated observation $\mathbf{o}$ formulated using the messages communicated by each agent. The agents communicate their encoded images, sensor measurements, and the messages received from other agents to the central critic. Therefore, for our two-agent system, the concatenated observation available to the central critic is 
\begin{equation*}
    \mathbf{o}: = (o^{(1)},o^{(2)}, f(v^{(1)};\xi_{e}^{(1)}), f(v^{(2)};\xi_{e}^{(2)}), m^{(1)}, m^{(2)}),
\end{equation*}
where $o^{(i)}$, $i \in \{1,2\}$, denotes the $i^{\text{th}}$ agent's sensor measurements,  $f(v^{(i)};\xi_{e}^{(i)})$ denotes the encoded image from the $i$\textsuperscript{th} agent, $m^{(i)}$ denotes the messages received by the $i$\textsuperscript{th} agent, $v^{(1)}$ denotes the first agent's  stacked frame image in Figure~\ref{fig: image1}, and $v^{(2)}$ denotes the second agent's color image, as in Figure~\ref{fig: image2}.

\begin{figure}[h!]
\centering
 \includegraphics[width=1.0\textwidth]{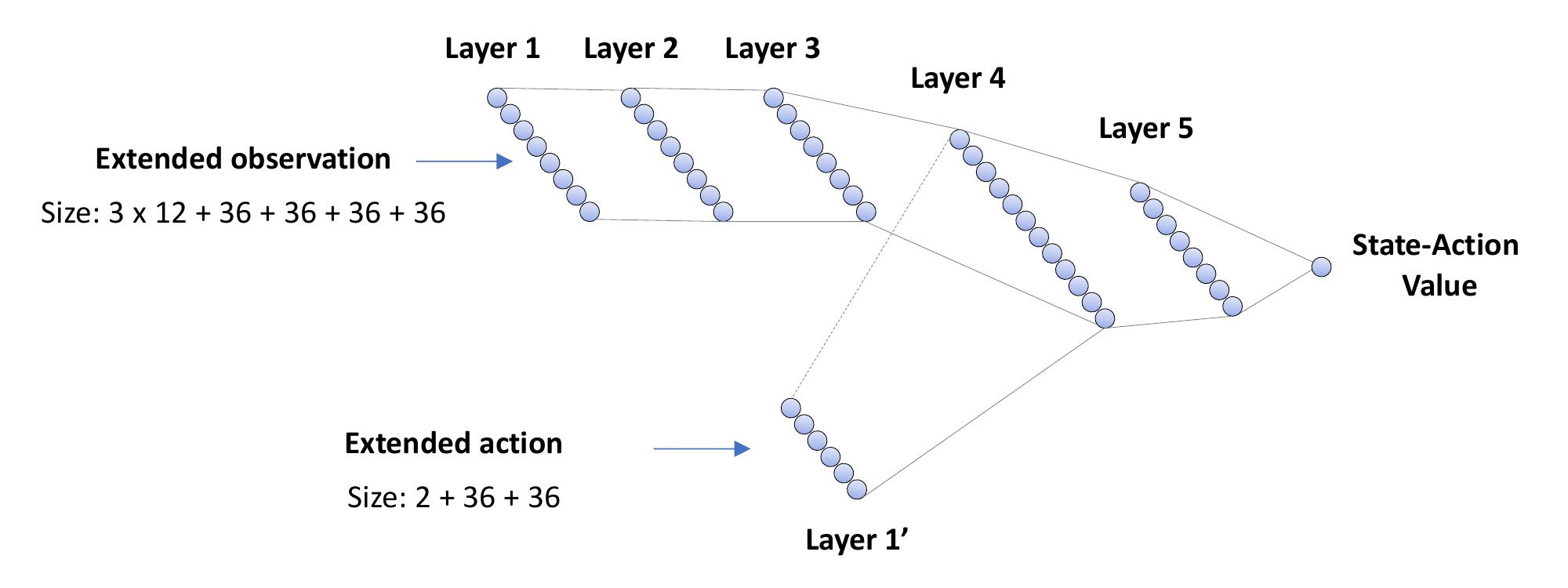}
 \caption{The computation graph of the critic-network that uses the extended observation, which is the concatenation of encoded images, the received messages and the vector observations.}
\medskip
\label{fig: critic network computation graph}
\end{figure}

The first agent's actor selects an action $a^{(1)}_{m}$ to \emph{actuate} the agent and an action $a^{(1)}_{c}$ to \emph{communicate} based on the local observations and received messages. At the next time-step, the communicated messages update the received messages, i.e. $m^{(2)} \leftarrow a^{(1)}_{c}$ and $m^{(1)} \leftarrow a^{(2)}_{c}$. As defined in~\eqref{eqn:action}, the extended action by the first agent is
\begin{equation*}\label{eq: ext_act_agent}
    \mathbf{a}^{(1)}:= (a^{(1)}_{m}, a^{(1)}_{c}),
\end{equation*} where we have dropped the temporal dependence.
We also define the extended abstract observation for the first agent as
\begin{equation*}\label{eq: ext_obs_agent}
    \mathbf{o}^{(1)}:= (o^{(1)},\, f(v^{(1)};\xi_{e}^{(1)}), \, m^{(1)}).
\end{equation*}
Similarly, we define the extended action $\mathbf{a}^{(2)}:= (a^{(2)}_{m}, a^{(2)}_{c})$ and observation $\mathbf{o}^{(2)}:= (o^{(2)}, f(v^{(2)};\xi_{e}^{(2)}), m^{(2)})$ for the second agent. The extended abstract observation $\mathbf{o}=(\mathbf{o}^{(1)},\mathbf{o}^{(2)} )$ and the extended action $(\mathbf{a}^{(1)}, \mathbf{a}^{(2)})$ are the inputs to the central critic network, as illustrated in Figure~\ref{fig: critic network computation graph}, and are used to calculate the state-action value. 

\begin{algorithm}[thpb]
  \caption{The Extended MADDPG for the Multi-Agent RL illustrated in Figure~\ref{fig: RL_Envrionment}.}\label{euclid}
  \begin{algorithmic}[1]
  \For{\texttt{episode=1 to M}}
        \State Choose the step size function: $\epsilon_1(t)$, $\epsilon_2(t)$, $\epsilon_3(t)$.
        \State Initialize parameters: $\theta^{(1)}$, $\theta^{(2)}$, $\theta^{(1)}_{\text{target}}$, $\theta^{(2)}_\text{target}$, $w$, $w_\text{target}$, \,$\xi_e^{(1)}$, $\xi_d^{(1)}$, $\xi_e^{(2)}$, $\xi_d^{(2)}$.
        
        \State Initialize a random process $\mathcal{N}_t$, for actuator action exploration.
        \State {\tt Central critic}: Receive initial abstract observation $\mathbf{o} \leftarrow (o^{(1)},o^{(2)}, f(v^{(1)};\xi_{e}^{(1)}), f(v^{(2)};\xi_{e}^{(2)}), m^{(1)}, m^{(2)})$ using encoders $f(\cdot;\xi_{e}^{(i)})$.
        
        \For{\texttt{t=1 to max-episode-length}}
                \For{\texttt{agent i= 1,2}}
                    \State Select the extended action $\mathbf{a}^{(i)} = \mu_{\theta^{(i)}}^{(i)}(\mathbf{o}^{(i)})$.
                    \State Execute exploratory action i.e. $a^{(i)}_{m}+\mathcal{N}_t$ on the environment and receive reward $r$.
                    \State Collect the new sensor observation $o^{(i)}, v^{(i)}$ on the state of the environment.
                    
            \State Locally store the raw images $(v^{(i)})$ in the image buffer $\mathcal{D}_{\text{image}}^{(i)}$.
            \State {\bf Local encoder update}: 
            \State Sample a random mini-batch $\mathcal{M}_{\text{image}}^{(i)}$ from $\mathcal{D}_{\text{image}}^{(i)}$. 
            \State Update the actor-encoders using gradient descent with the step-size $\epsilon_3(t)$ to minimize the loss,
            \begin{equation*}
                l(\xi_{e}^{(i)},\xi_{d}^{(i)}) = \frac{1}{M}\sum_{m=1}^M d(v_m, g(f(v_m;\xi_{e}^{(i)});\xi_{d}^{(i)})), \quad v_m \in \mathcal{M}_\text{image}^{(i)}.
            \end{equation*}
            \State {\bf Local actor update}: 
            \State Receive the sampled mini-batch $\mathcal{M}$ from the central critic.
            \State Update the $i$\textsuperscript{st} actor with the step-size $\epsilon_2(t)$ using the sampled policy gradient:
            \begin{equation*}
            \nabla_{\theta^{(i)}}J 	\approx  \frac{1}{M}\sum_{m=1}^M \nabla_{\mathbf{a}^{(i)}}Q(\mathbf{o}_m,\mathbf{a}^{(i)}, \mathbf{a}_m^{(i-)} ;w)|_{\mathbf{a}^{(i)}=\mu^{(i)}(\mathbf{o}_m)}  \nabla_{\theta^{(i)}} \mu_{\theta^{(i)}}^{(i)} (\mathbf{o}_m),
            \end{equation*}
            \State where $i- = 2$ if $i=1$ and vice versa.
             \State Update the parameters of the target networks $\theta^{(i)}_\text{target} \leftarrow \tau \theta^{(i)} + (1-\tau)\theta^{(i)}_\text{target}$.
            
               \State {\bf Communication}: 
               \State $m^{(i-)} \leftarrow a^{(i)}_{c}$ and send it to $(i-)$th agent. Receive message $m^{(i)} \leftarrow a^{(i-)}_{c}$ from $(i-)$th agent.
 
                \EndFor

            \For{\texttt{Central Critic}}
            
            \State Receive $(o^{(1)},o^{(2)}, f(v^{(1)};\xi_{e}^{(1)}), f(v^{(2)};\xi_{e}^{(2)}),\, m^{(1)}, m^{(2)})$ from the local agents.
                \State $\mathbf{o}' \rightarrow (o^{(1)},o^{(2)}, f(v^{(1)};\xi_{e}^{(1)}), f(v^{(2)};\xi_{e}^{(2)}),\, m^{(1)}, m^{(2)})$.
                \State Store the transition $(\mathbf{o}, \mathbf{a}^{(1)}, \mathbf{a}^{(2)}, r, \mathbf{o}')$ in replay buffer $\mathcal{D}_\text{replay}$. 
                \State Sample a random mini-batch $\mathcal{M}$ from $\mathcal{D}_\text{replay}$
            \begin{equation*}
                \mathcal{M}=(\,(\mathbf{o}_{1}, \mathbf{a}^{(1)}_{1}, \mathbf{a}^{(2)}_{1}, r_{1}, {\mathbf{o}'}_{1})\,, \dots\,, (\mathbf{o}_{M}, \mathbf{a}^{(1)}_{M}, \mathbf{a}^{(2)}_{M}, r_{M}, {\mathbf{o}'}_{M})\,).
            \end{equation*}
            \State Send $\mathcal{M}$ and $w$ to all local agents.
            \State Set $y_m = r_m + \gamma Q(\mathbf{o}'_m, \mathbf{a}^{(1)}, \mathbf{a}^{(2)}; w_\text{target})|_{\mathbf{a}^{(1)} = \mu^{(1)}(\mathbf{o}'_m;\theta^{(1)}_\text{target}),\mathbf{a}^{(2)} = \mu^{(2)}(\mathbf{o}'_m;\theta^{(2)}_\text{target})}$.
            \State Update the central critic parameter $w$ with the step-size $\epsilon_1(t)$ by minimizing the loss,
            \begin{equation*}
                l(w)=\frac{1}{M}\sum_{m=1}^M(Q(\mathbf{o}_m, \mathbf{a}^{(1)}_m, \mathbf{a}^{(2)}_m;w )-y_m)^2.
            \end{equation*}
            
                        \State Update the parameters of the target networks $w_\text{target} \leftarrow \tau w + (1-\tau)w_\text{target}$.

            \EndFor

            \State $\mathbf{o} \leftarrow \mathbf{o}'$.

        \EndFor
  \EndFor
  
  \end{algorithmic}
  \label{alg1}
\end{algorithm}
\newpage

\section{Simulation Results}
\begin{figure}[h!]
\centering
\begin{subfigure}{.5\textwidth}
  \centering
  \includegraphics[width=1.\linewidth]{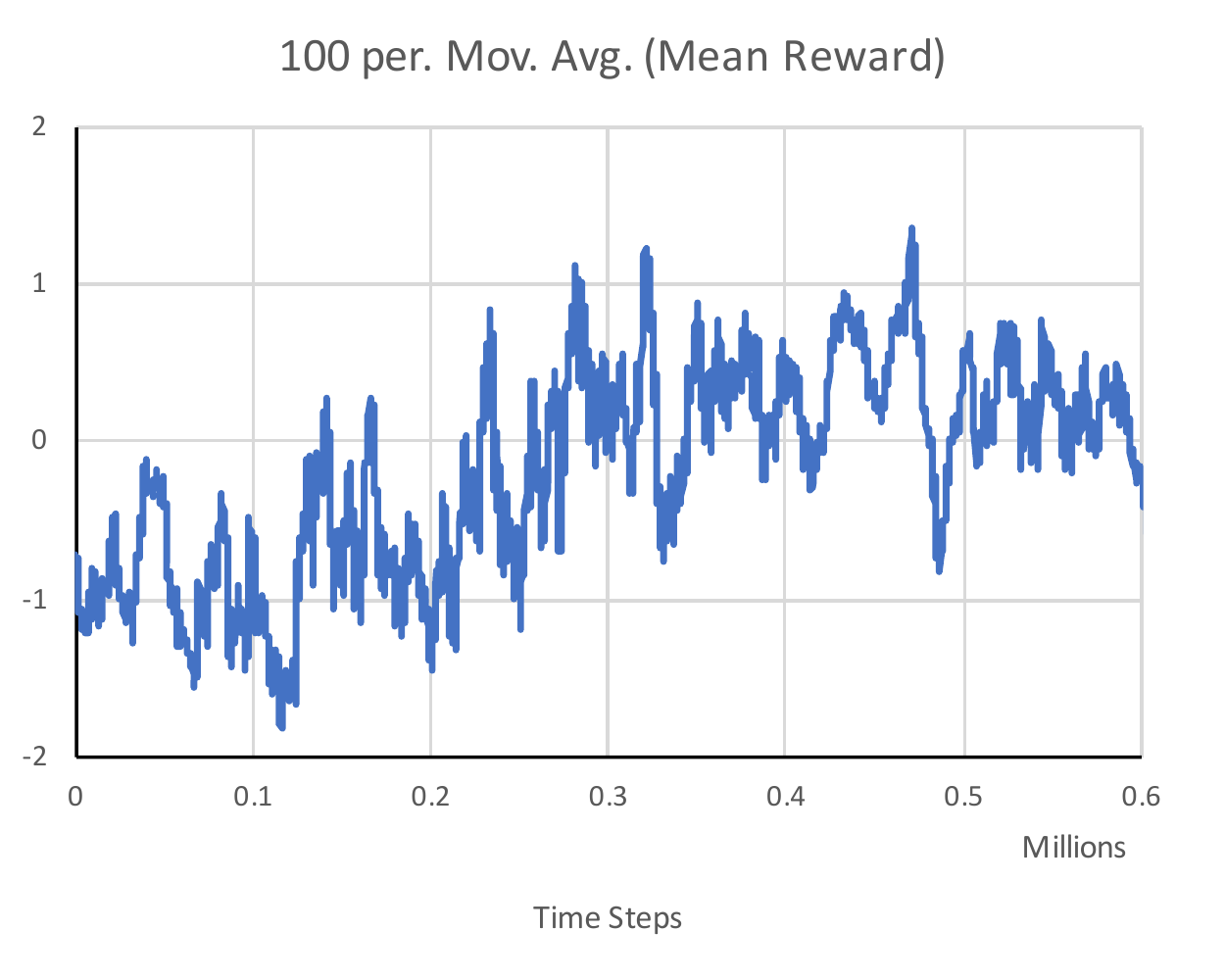}
  \caption{Moving average of the mean rewards.}
  \label{chart1}
\end{subfigure}%
\begin{subfigure}{.5\textwidth}
  \centering
  \includegraphics[width=1.\linewidth]{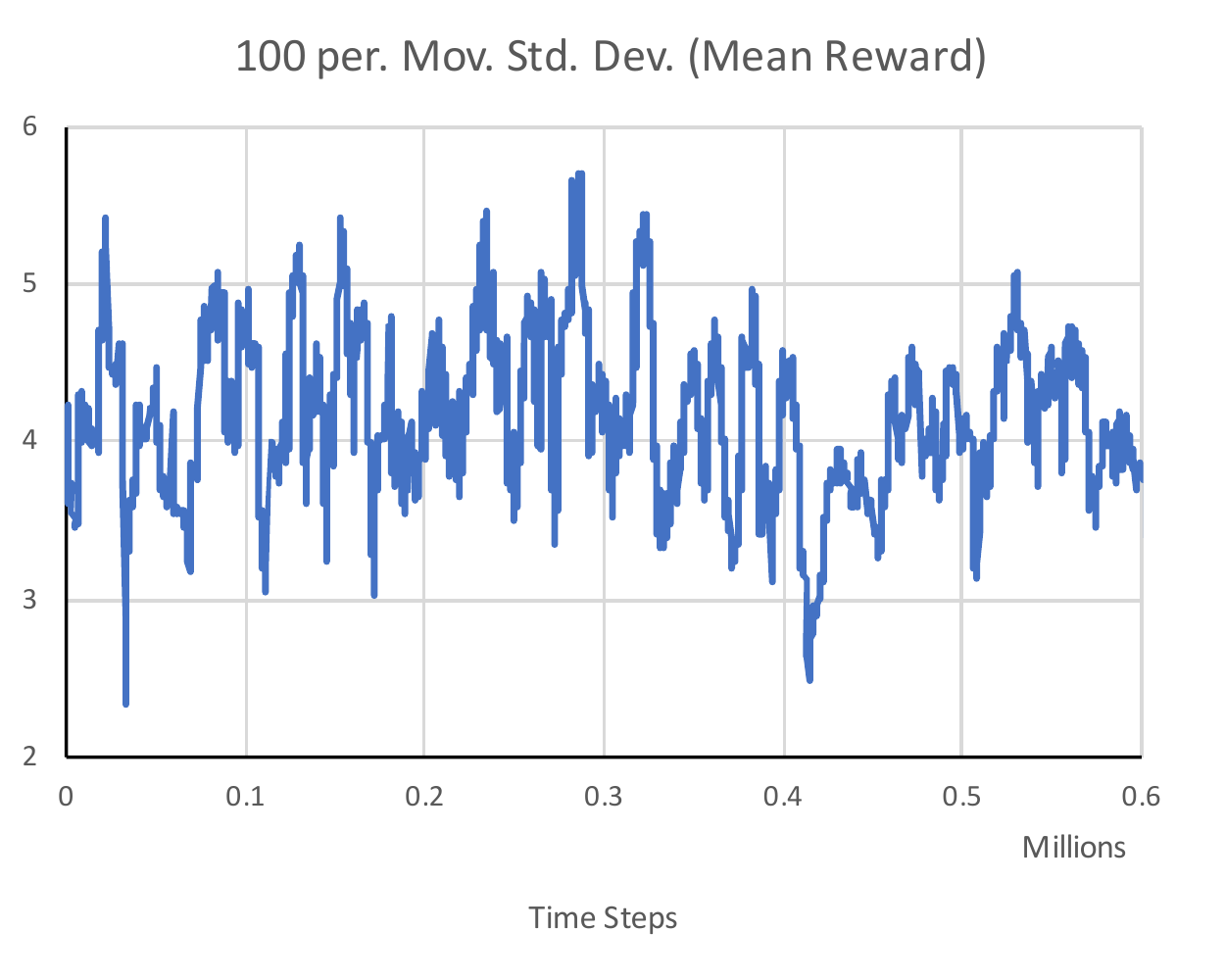}
  \caption{Moving standard deviation of the mean rewards.}
  \label{chart2}
\end{subfigure}
\caption{Moving average and standard deviation of the mean rewards per episode.}
\label{fig: likelihood_and_sigma}
\end{figure}

\begin{figure}[h!]
\centering
\begin{subfigure}{.5\textwidth}
  \centering
  \includegraphics[width=1.\linewidth]{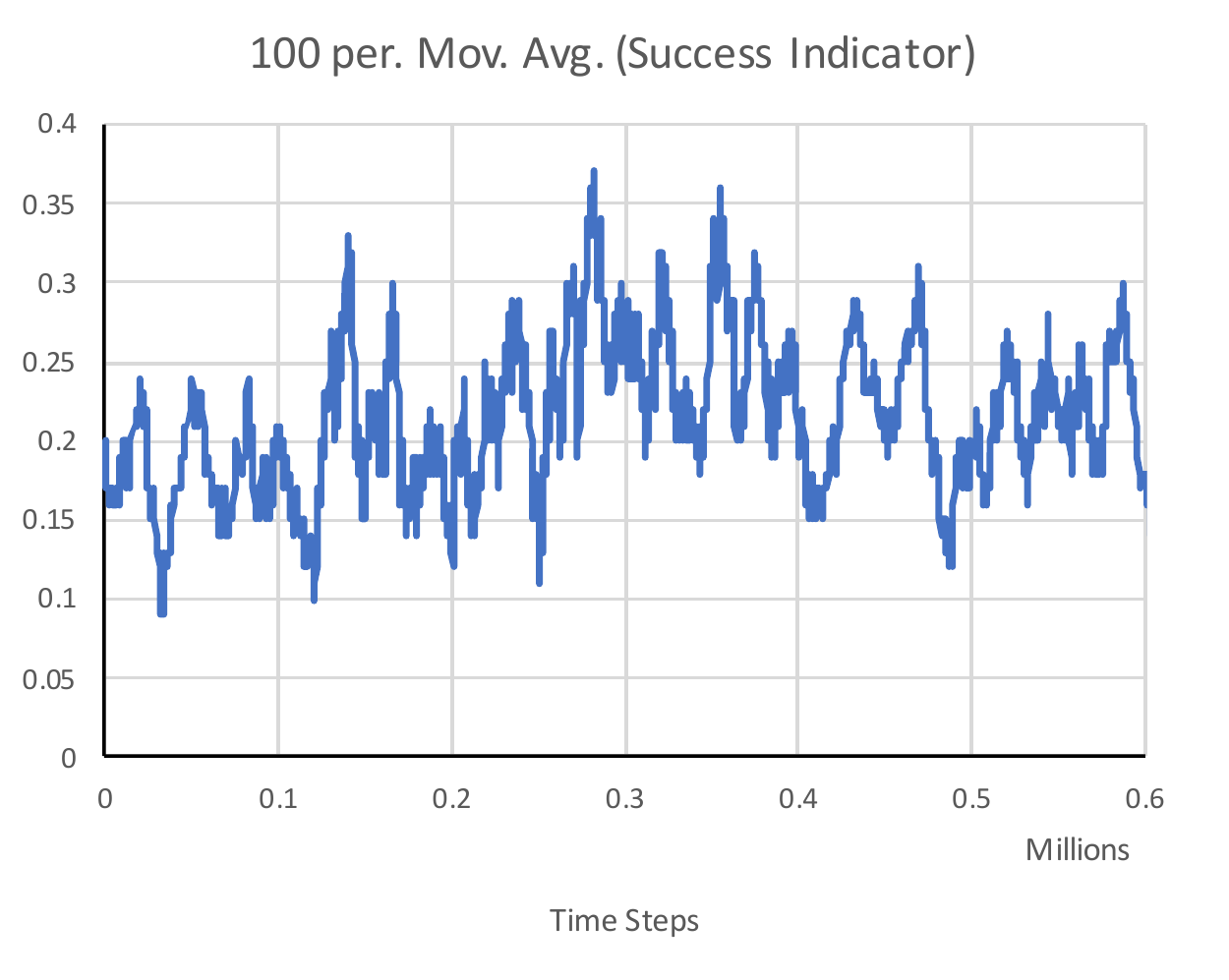}
  \caption{Moving average of the success indicator.}
  \label{chart3}
\end{subfigure}%
\begin{subfigure}{.5\textwidth}
  \centering
  \includegraphics[width=1.\linewidth]{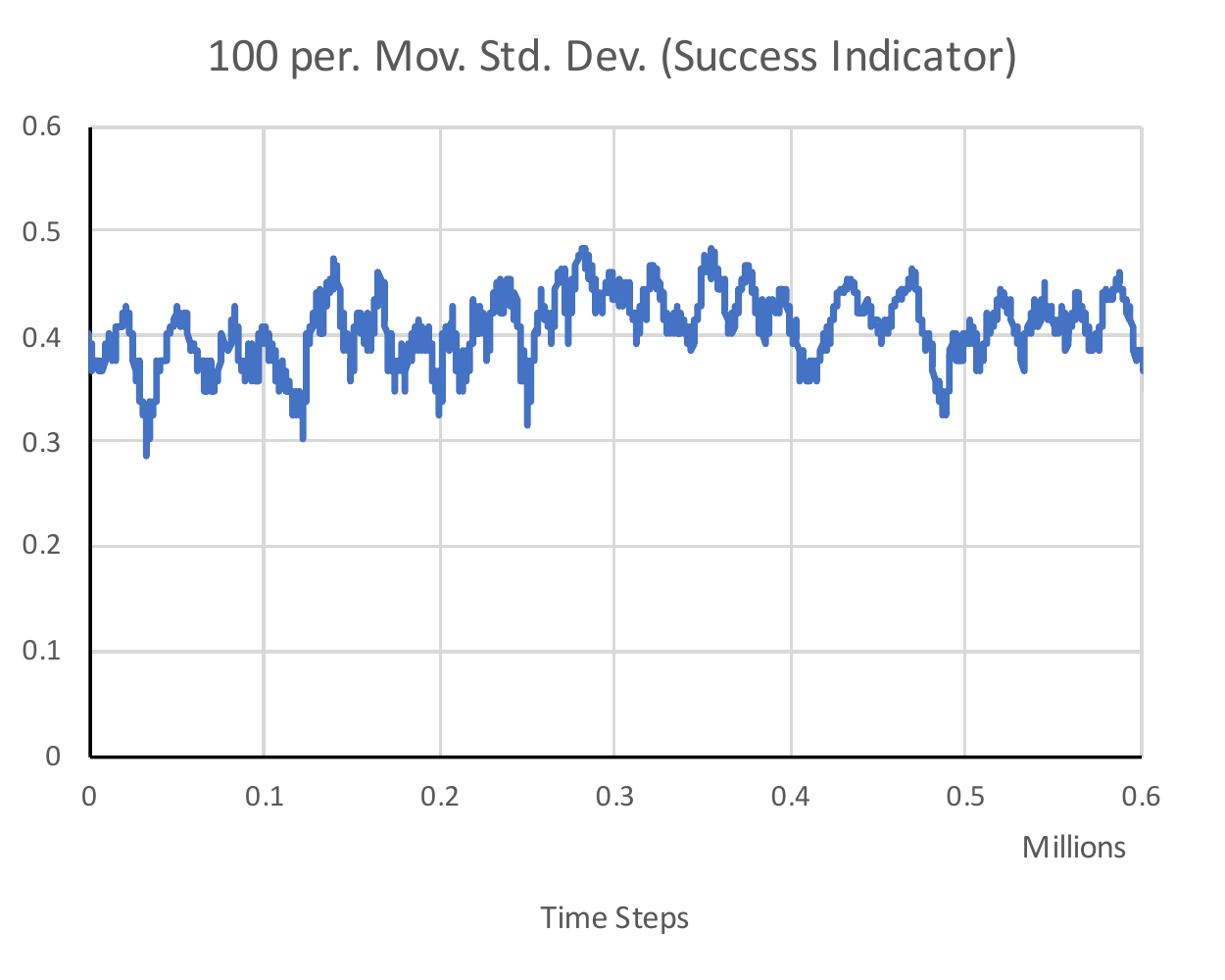}
  \caption{Moving standard deviation of the success indicator.}
  \label{chart4}
\end{subfigure}
\caption{Success indicator (1: successful episode, 0: otherwise).}
\label{fig: success indicator}
\end{figure}

We ran one million simulation steps to verify the feasibility of the proposed algorithm. The reinforcement learning algorithm improves the mean rewards per episode, as shown in Figure~\ref{chart1}. Also, the rate of the successful episodes has been improved, as shown in Figure~\ref{chart3}. However, the success indicator values in Figure~\ref{chart3} show that only 1 out of 4 trials is successful. There are a few possible reasons why the performance of the trained policy is not satisfactory. First, the incomplete state observation is a fundamental challenge for the Markov assumption based reinforcement algorithms. Our proposed algorithm assumes that the autoencoder's image compression preserves useful information for the task. However, the loss function for the autoencoder in~\eqref{eq: encoder_loss} only considers the mean squared error between the raw image and the reconstructed image. Without incorporating the reward of the task in the training of the autoencoder might result in the loss of the state information, which is necessary for the successful task completion. 

\begin{figure}[h!]
\centering
 \includegraphics[width=0.6\textwidth]{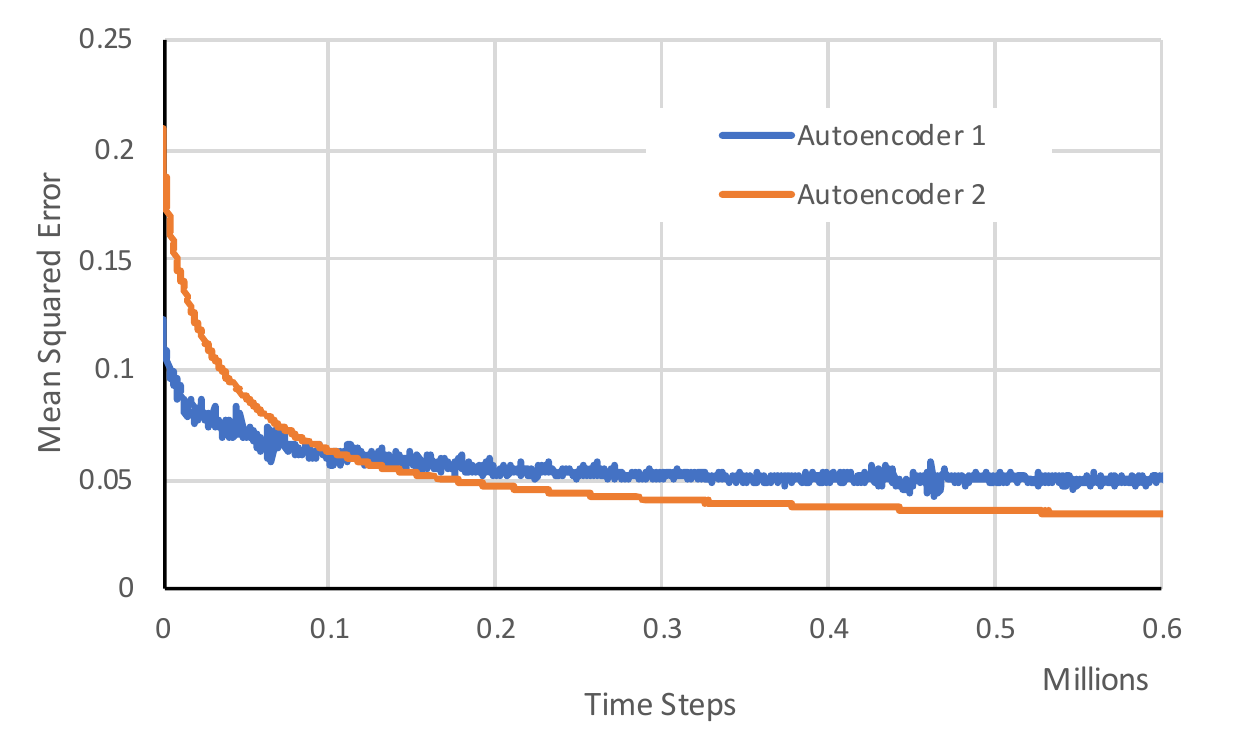}
 \caption{Mean squared error (MSE) between the raw images and the images reconstructed by the autoencoders. Autoencoder 1 processes the stacked frames in Figure~\ref{fig: image1} and Autoencoder 2 processes the color images in Figure~\ref{fig: image2}.}
\medskip
\label{fig: autoencoder error}
\end{figure}

\begin{figure}[h!]
\centering
 \includegraphics[width=0.6\textwidth]{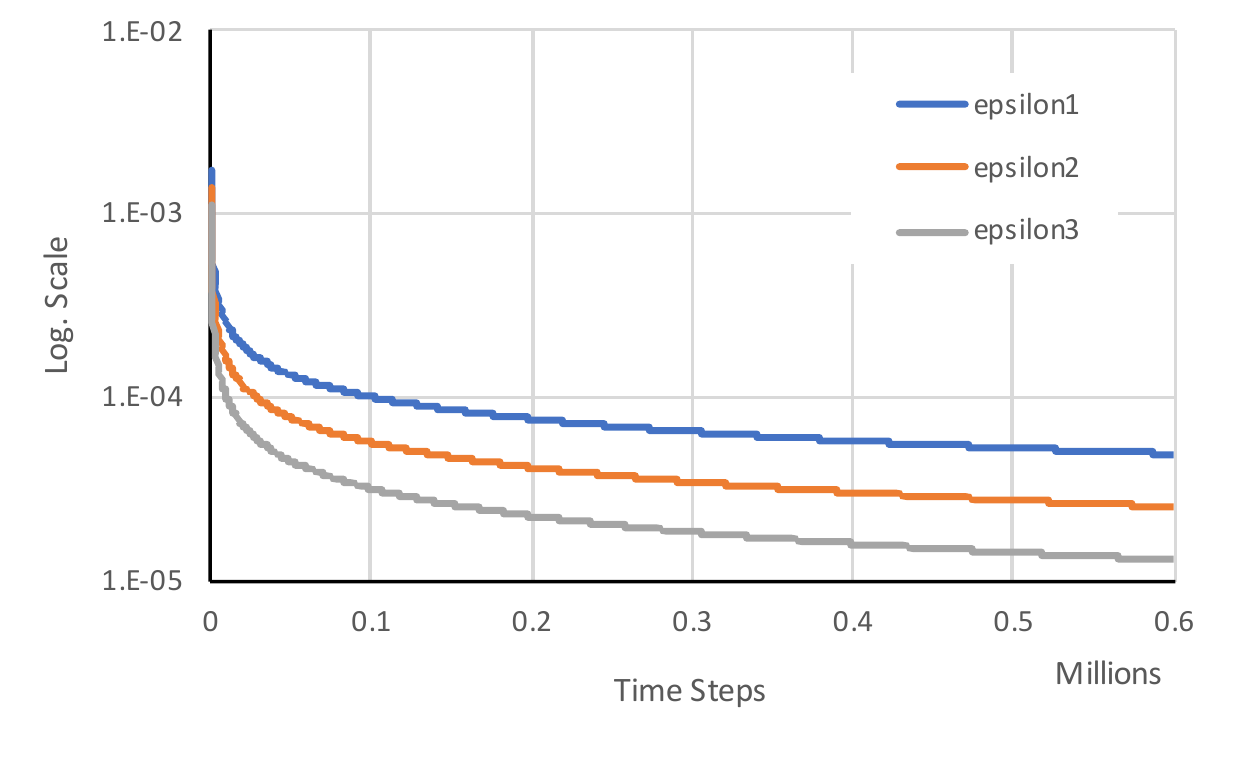}
 \caption{The optimization scheduler decays the step-sizes to update the parameters. The fastest step-size $\epsilon_1(t)$ is for the central critic update, slower step-size $\epsilon_2(t)$ is for the actors , and the slowest $\epsilon_3(t)$ is for the encoder.}
\medskip
\label{fig: step-size}
\end{figure}

In addition to the incomplete state observation, another challenge would be the uniform random initialization of an episode. Figure~\ref{chart2} shows large variations of mean reward per episode despite decaying learning rate. As the training progresses, the step-size (or learning rate) is decreased as shown in Figure~\ref{fig: step-size} so that the actors (control policy) eventually stop evolving. However, there are still large variations in the mean reward. This suggests that the variation is mostly due to the uniformly random initialization, since the policy itself is being very slowly updated in the later training iterations. Obviously, robustness to the random initialization is desired for the reinforcement learning algorithm. However, the uniform randomness here might not be appropriate, since it makes  difficult for the reinforcement learning algorithm to recognize patterns. Also, the environments (Atari games, robot dynamics, etc.) used in the successful deep reinforcement learning papers~\cite{mnih2015human, lillicrap2015continuous} are more deterministic regarding the initialization compared to our simulation environment.
%\newpage

\section{Conclusion}
We present a reinforcement learning framework that takes into account decision making regarding sharing local information between agents for collaborative tasks. We use autoencoders to reduce the dimensions of the camera images taken by each agent. The encoded visual observations are fed to an actor-network, which decides communication messages to the other agents. The data compression by local agents makes it possible to implement deep reinforcement learning algorithms~\cite{mnih2015human, lillicrap2015continuous}, which use a large replay buffer ($10^6$ samples). To validate the feasibility of the proposed framework, we develop a 3D simulation environment that provides camera images from different perspectives of the two heterogeneous agents. Simulation results show that the algorithm improves the performance of a given collaborative task.

\subsection{Future Work}
The proposed idea to use autoencoders to compress local visual observations in order to send it to the central critic, while keeping the usage of the communication resource economically feasible, is convincing. In the algorithm, each agent updates the autoencoder independently of the other agent. Minimizing the mean squared error of the autoencoder without considering the collaborative task might not lead to useful image encoding. In order to learn the dependence between local visual observations and the collaborative task reward, the agents need to share the raw high dimensional observations. However, sharing the high-dimensional data over the network cannot be as frequent as sharing the low-dimensional compressed image, due to communication resource constraints. So the data sharing will be asynchronous and in multiple time-scales, depending on the types of data being shared. Multiple time-scales and the asynchronous stochastic algorithm from~\cite{kushner2003stochastic} have the potential to address this issue. Intuitively, communication protocols should not change drastically for the agents to perform collaborative tasks. Hence, the use of communication networks to send the high dimensional local visual observation to the central critic, in order to update the communication protocol (autoencoder), should be less frequent than the update of the local policy for each agent.

Furthermore, navigation tasks in the 3D environment require the reinforcement learning algorithm to memorize the previously taken path. Humans navigate by constructing a map while memorizing the previous path. In other words, the human estimates the global state (location of the human in the map) based on the previous trajectory (taken path). So the 3D navigation is inherently a partially observable Markov decision process (POMDP), since the global state (the location) is hidden. To address the challenge of the POMDP, the use of the internal memory to predict or infer hidden states in the context of the reinforcement learning task was proposed in~\cite{heess2015memory, hausknecht2015deep}. For the 3D game environment, a recurrent neural network was employed to overcome the inherent partial observability of the navigation task~\cite{lample2017playing}. However, the training of the recurrent neural network in~\cite{heess2015memory, hausknecht2015deep,lample2017playing} only uses the Bellman optimality equation, which does not consider how well the recurrent neural network can predict or infer hidden states. Maximum likelihood estimate (MLE) framework is a natural approach to infer hidden states. In~\cite{yoon2018hidden}, the authors proposed an on-line HMM estimation based reinforcement learning algorithm, which converges to the maximum likelihood estimate.

Advantages of using autoencoders or recurrent neural networks in reinforcement learning tasks can be from their capability to mimic the true environment, in the sense that the learning algorithm can predict or infer hidden states. The future work will focus on devising algorithms, which converge to the maximum likelihood estimate of the environment model that uses the autoencoder and the recurrent neural network. The estimated environment model will be used to construct state estimator. Reinforcement learning algorithms with the state estimator will be applied to various collaborative tasks in the 3D simulation environment.

Deploying the vision-based reinforcement learning to real cooperative missions as the ones in~\cite{kaminer2017time} has the potential to improve the resilience of  multiagent systems to communication failures. For example, camera images contain the information on the location of other agents, which can be extrapolated to achieve coordination in the presence of communication failures. However, training of such intelligent agents requires significant random exploration to determine the optimal policies. Therefore, a method to {\em safely explore the policy space} is required to deploy the reinforcement learning algorithms on real missions. The future work will investigate methods to autonomously constrain the policy space, while learning the optimal policy.

\section*{Acknowledgment}
This material is based upon work supported by the National Science Foundation under National Robotics Initiative grant \#1830639 and Air Force Office of Scientific Research grant \#FA9550-18-1-0269.
% The NSF grant number 1830639 is for NRI: INT: COLLAB: Synergetic Drone Delivery Network in Metropolis. 

% Is there a grant number that I can add for AFOSR ?

\bibliography{main}

\end{document}